%% file: HT_nmin.tex
\documentclass[10pt, conference, compsocconf]{IEEEtran}
\IEEEoverridecommandlockouts
\usepackage{cite}
\usepackage{amsmath,amssymb,amsfonts}
\usepackage{algorithm}
\usepackage{algorithmic}
\usepackage{graphicx}
\usepackage{textcomp}
\usepackage{booktabs} 
\usepackage{enumerate}
\usepackage[utf8]{inputenc}

\usepackage{xcolor} 
\usepackage{mathtools}
\DeclarePairedDelimiter{\ceil}{\lceil}{\rceil}
\usepackage[hyphens]{url}

\graphicspath{{./figures/}}
\usepackage{url}

\usepackage{xcolor}
\def\BibTeX{{\rm B\kern-.05em{\sc i\kern-.025em b}\kern-.08em
    T\kern-.1667em\lower.7ex\hbox{E}\kern-.125emX}}
\begin{document}

\title{Hoeffding Trees with \textit{nmin adaptation}
}


\author{
\IEEEauthorblockN{Eva Garc\'ia-Mart\'in\IEEEauthorrefmark{1}, 
	Niklas Lavesson\IEEEauthorrefmark{1}\textsuperscript{,}\IEEEauthorrefmark{2},
	H\aa kan Grahn\IEEEauthorrefmark{1}, 
	Emiliano Casalicchio\IEEEauthorrefmark{1}\textsuperscript{,}\IEEEauthorrefmark{3} and 
	Veselka Boeva\IEEEauthorrefmark{1}} 
	\IEEEauthorblockA{
	\IEEEauthorrefmark{1} Blekinge Institute of Technology, Karlskrona, Sweden \\
	\IEEEauthorrefmark{2} Jönköping University, Jönköping, Sweden \\
	\IEEEauthorrefmark{3} Sapienza University of Rome, Rome, Italy \\
	Email: \{eva.garcia.martin, niklas.lavesson, hakan.grahn, emiliano.casalicchio, veselka.boeva\}@bth.se \\
	}
}







\maketitle

\begin{abstract}
Machine learning software accounts for a significant amount of energy consumed in data centers. 
These algorithms are usually optimized towards predictive performance, i.e. accuracy, and scalability. 
This is the case of data stream mining algorithms.
Although these algorithms are adaptive to the incoming data, they have fixed parameters from the beginning of the execution. We have observed that having fixed parameters lead to unnecessary computations, thus making the algorithm energy inefficient. 

In this paper we present the \textit{nmin adaptation} method for Hoeffding trees. This method adapts the value of the \textit{nmin} parameter, which significantly affects the energy consumption of the algorithm.
The method reduces unnecessary computations and memory accesses, thus reducing the energy, while the accuracy is only marginally affected.
We experimentally compared VFDT (Very Fast Decision Tree, the first Hoeffding tree algorithm) and CVFDT (Concept-adapting VFDT) with the VFDT-\textit{nmin} (VFDT with \textit{nmin adaptation}).
The results show that VFDT-\textit{nmin} consumes up to 27\% less energy than the standard VFDT, and up to 92\% less energy than CVFDT, trading off a few percent of accuracy in a few datasets. 
\end{abstract}


\section{Introduction}
Large-scale data centers account for a significant share of the energy consumption in many countries~\cite{dataCentersEnergy}. The number of data centers and the computational demand is rapidly increasing due to the rate at which data is generated and processed. 
Although machine learning algorithms are responsible for some part of that computation, since they are introduced in almost all application domains, they are seldom optimized w.r.t. their energy consumption. 
State-of-the-art algorithms that can have an impact on energy consumption are data stream mining algorithms~\cite{gaber2012advances}, since they are designed to run continuously on embedded systems.


Although data stream mining algorithms adapt the decision model based on the incoming data, i.e. concept drift adaptation, the parameters of such algorithms are fixed from the beginning of the execution. 
We have observed that having fixed parameters leads to the algorithm making unnecessary computations, thus increasing its energy consumption~\cite{garcia2017identification}. 

In this paper we propose dynamic parameter adaptation, a method to reduce the energy consumption without sacrificing accuracy. 
We illustrate this with the \textit{nmin adaptation} method to improve parameter adaptation in Hoeffding trees. Hoeffding tree algorithms evaluate if $nmin$ instances observed at a node are enough to make a confident split. 
However, the $nmin$ parameter has a significant impact on the overall energy consumption of the VFDT, visible in its energy model (Section~\ref{sec:vfdt_energymodel}). Thus, having a fixed $nmin$ value that does not adapt to the incoming data leads to energy inefficiencies.
We propose \textit{nmin adaptation} to adapt the value of $nmin$ depending on the incoming data, to ensure that the algorithm calculates the best attributes only when there will be a split. This reduces the amount of computation related to calculating information gain of all the attributes, thus reducing its energy consumption. 
This method has the following properties: 
\begin{enumerate}[i)]
	\item Adaptive to the characteristics of the data
	\item Unique value of $nmin$ for each tree node
	\item Applicable to any Hoeffding Tree algorithm
\end{enumerate}

We experimentally compare the VFDT (Very Fast Decision Tree~\cite{domingos2000mining}, the first Hoeffding tree algorithm), with the VFDT-\textit{nmin} (VFDT with \textit{nmin adaptation}), and CVFDT (Concept-Adapting Very Fast Decision Tree~\cite{hulten2001mining}) on 15 datasets. The results show that VFDT-\textit{nmin} reduces the energy consumption significantly in comparison to VFDT and CVFDT, yielding an average of 9.5\% and up to 27\% energy reduction compared to the VFDT, and an average of 85\% energy reduction in comparison to the CVFDT. The predictive performance, i.e. the accuracy, is only decreased slightly by this energy reduction (less than 1\% average loss for VFDT-\textit{nmin} in comparison to VFDT). 

The paper is organized as follows. 
The background and related work are presented in Section~\ref{sec:background}. 
The \textit{nmin adaptation} method and main contribution of this paper is presented in Section~\ref{sec:nmin_adapt}.
Section~\ref{sec:energy_cons_vfdt} profiles the energy consumption of the VFDT presenting a theoretical energy model for such algorithm. 
Section~\ref{sec:experimental_design} presents the experimental design.
Sections~\ref{sec:results} and \ref{sec:discussion} present the results and discussion.
Section~\ref{sec:conclusions} concludes the paper with the significance and impact of our work.

\section {Background and Related Work}
\label{sec:background}

This section explains the fundamentals of the VFDT, and finishes by explaining related studies in streaming data, green computing, and resource-aware machine learning.


\subsection{VFDT}
\label{sec:vfdt}

Very Fast Decision Tree~\cite{domingos2000mining} is a decision tree algorithm that builds a tree incrementally. The data instances are analyzed sequentially and only once. The algorithm reads an instance, sorts it into the corresponding leaf, and updates the statistics at that leaf. To update the statistics the algorithm maintains a table for each node, with the observed attribute and class values. Updating the statistics of numerical attributes is done by saving and updating the mean and standard deviation for every new instance. 
Each leaf also stores the instances observed so far. After every $nmin$ read instances at that leaf, the algorithm calculates the information gain ($\overline{G}$) from all observed attributes. The difference in information gain between the best and the second best attribute ($\Delta \overline{G}$) is compared with the Hoeffding Bound~\cite{hoeffding1963probability} ($\epsilon$). If $\Delta \overline{G} > \epsilon$, then that leaf is substituted by a node, and there is a split on the best attribute. That attribute is removed from the list of attributes available to split in that branch.  
If $\Delta \overline{G} < \epsilon < \tau$, a tie occurs, splitting on any of the two top attributes, since they have very similar information gain values. 
The Hoeffding Bound ($\epsilon$), 
\begin{equation} \label{eq:hoeff}
\epsilon = \sqrt{ \frac{ R^2 \ln (1/\delta)}{ 2n}} 
\end{equation}
states that the chosen attribute at a specific node after seeing $n$ number of examples, will be the same attribute as if the algorithm has seen an infinite number of examples, with probability $1-\delta$.

We now discuss the computational complexity of the VFDT, shown in lines 1-21 from Alg.~\ref{alg:vfdt-nmin}. Suppose that $n$ is the number of instances and $m$ is the number of attributes. The algorithm is a loop over $n$ iterations. Every step between 6 and 9 require execution time that is  proportional to $m$. In the worst case scenario the computational complexity of step 7 is $O(m)$ according to~\cite{domingos2000mining}. The function in step 7 traverses the tree until it finds the corresponding leaf. Since the attributes are not repeated for each branch, in the worst case scenario the tree will have a depth of $m$ attributes. 
Step 8 runs in constant time. The computational complexity of this part can be evaluated to $O(n \cdot m)$. The computational complexity of the remainder part of the algorithm (from step 11 downwards) depends on $n/nmin$. Moreover, the computational complexity of steps 11 to 13 is equal to $O(m)$, while steps 16 to 18 need constant time, i.e. the computational complexity of this part is $O(n/nmin \cdot m)$. The total computational complexity of the VFDT is $O(n \cdot m) + O(n/nmin \cdot m) $ and $n>> nmin$, i.e. it can be simplified to $O(n \cdot m)$.

\subsection{Related Work}
Energy efficiency is an important research topic in computer engineering~\cite{dubois2012parallel}. 
Reams~\cite{reams2012modelling} provides a good overview of energy-efficiency in computing for different platforms:  servers, desktops and mobile devices. The author also proposes an energy cost model based on the number of instructions, power consumption, the price per unit of energy, and the execution time. 
While energy efficiency has mostly been studied in computer engineering, during the past years green computing has emerged. 
Green IT, also known as green computing, started in 1992 with the launch of the Energy Star program by the US Environmental Protection Agency (EPA)~\cite{ruth2009green}.
Green computing is \textit{the study and practice of designing, manufacturing, using, and disposing computers, servers, and associated systems efficiently and effective with minimal or no environmental impact}~\cite{ruth2009green}. One specific area is energy-efficient computing~\cite{reams2012modelling}, where there is a significant focus on reducing the energy consumption of data centers~\cite{shehabi2016united}. 



In relation to big data, data centers, and cloud computing, there have been several studies that design methods for energy-efficient cloud computing~\cite{rhoden2011improving,chen2012energy}. One approach was used by Google Deep Mind to reduce the energy used in cooling their data centers~\cite{deepmindenergy}. 
These studies focus on reducing the energy consumed by data centers using machine learning to, e.g., predict the load for optimization. However, we focus on reducing the energy consumption of machine learning algorithms.




Regarding machine learning and energy efficiency, there has been a recent surge of interest towards resource-aware machine learning.
The focus has been on building energy efficient algorithms that are able to run on platforms with scarce resources~\cite{gaber2005board,Gaber2004,dortmund,kotthaus/2016a}.
Closely related is the work done on building energy-efficient deep neural networks~\cite{yang2016designing}. They develop a model where the energy cost of the principal components of a neural network is defined, and then used for pruning a neural network without reducing accuracy.


Data stream mining algorithms analyze data aiming at reducing the memory usage, by reading the data only once without storing it.
Examples of efficient algorithms are the VFDT~\cite{domingos2000mining} and a KNN streaming version with self-adjusting memory~\cite{losing2016knn}. 
There have been extensions to these algorithms for distributed systems, such as the Vertical Hoeffding Tree~\cite{kourtellis2016vht}, where the authors parallelize the induction of Hoeffding Trees; and the Streaming Parallel Decision Tree algorithm (SPDT). More focused on hardware approaches to improve Hoeffding trees is the work proposed by~\cite{marron2017low}, where they parallelize the execution of random forest of Hoeffding trees, together with a specific hardware configuration to improve induction of Hoeffding trees.
Other work has been done where the authors present the energy hotspots of the VFDT~\cite{garcia2017identification}. 
Our proposed work in this paper focuses on a direct approach to reduce the energy consumption of the VFDT by dynamically adapting the $nmin$ parameter based on incoming data, introducing the notion of \textit{dynamic parameter adaptation} in data stream mining.

\section{\textit{nmin} adaptation}
\label{sec:nmin_adapt}
The \textit{nmin adaptation} method, the main contribution of this paper, aims at reducing the energy consumption of the VFDT while maintaining similar levels of accuracy. 
There are many computations and memory accesses dependent on the parameter $nmin$, observed in the energy model presented in Section~\ref{sec:energy_cons_vfdt}.
However, the design of the original VFDT sets the value of $nmin$ to a fixed value from the beginning of the execution. This is problematic, because there are many functions that would be computed unnecessarily if $nmin$ instances are not enough to make a confident split (e.g. $calc\_entropy$, $calc\_hoeff\_bound$, and $get\_best\_att$).
Our goal is to set $nmin$ to the specific value that ensures a split, so that the $\frac{N}{nmin}$ values in \eqref{eq:VFDT_finalmodel} are only computed when needed. 
\textit{nmin adaptation} adapts the value of $nmin$ to a higher one, thus making $\frac{N}{nmin}$ smaller. This approach reduces computations, reduces memory accesses, and doesn't affect the final accuracy, since we are only computing those functions when needed. 

In another publication, the authors~\cite{garcia2017identification} already confirmed the high energy impact of the functions involved in calculating the best attributes. This matches with our energy model, and motivates the reasons and objectives for \textit{nmin adaptation}:
\begin{enumerate}
	\item Reduce the number of computations and memory accesses by adapting the value of $nmin$ to a specific value that ensures a split. 
	\item Maintain similar levels of accuracy by removing only unnecessary computations, thus developing the same tree structure. 
\end{enumerate}

\begin{figure}[!t]
  \centering
  \includegraphics[width=0.50\textwidth]{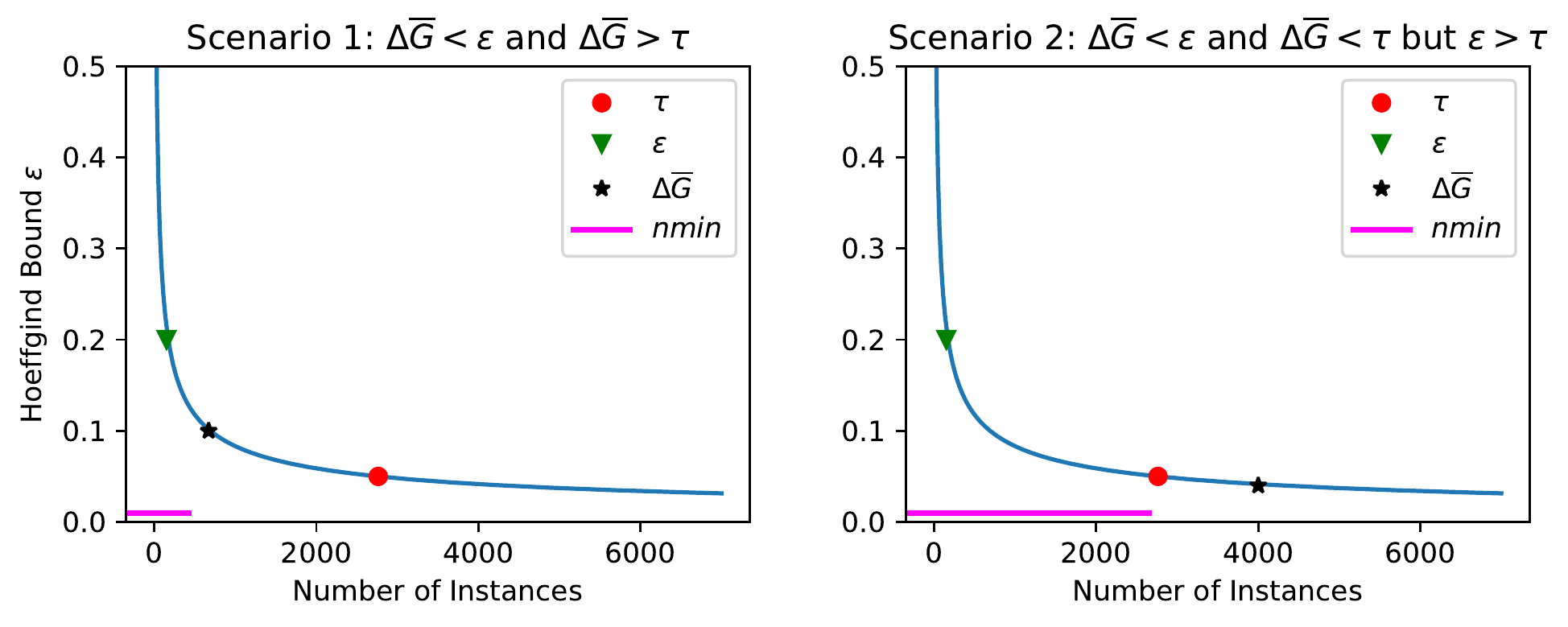}
  \caption{\small Variation of $\epsilon$ (Hoeffding bound) with the number of instances. \textit{nmin adaptation method} for scenarios 1 and 2. }
  \label{fig:icdm:epsilonvar}
  \end{figure}

\textit{nmin adaptation} sets $nmin$ to the estimated number of instances required to guarantee a split with confidence $1-\delta$.
  The higher the value of $nmin$, the higher the chance to split. However, setting $nmin$ to a high value decreases accuracy, and setting $nmin$ to a lower value increases the accuracy at the expense of energy, as it has to calculate the $\overline{G}$ of all attributes even when there are not enough instances to make a confident split.
  We have identified two scenarios that are responsible for not splitting. We set $nmin$ to a different value to address these scenarios, depending on the incoming data. This $nmin$ value is unique for every leaf, since different instances reach different leaves.
  This is a significant difference with the original VFDT, where they set the same $nmin$ for all leaves. 
  The two scenarios are the following:

\newtheorem{theorem}{Scenario}
\begin{theorem}[$\Delta \overline{G} < \epsilon$ and $\Delta \overline{G} > \tau$]
\label{scenar1o}
	Figure~\ref{fig:icdm:epsilonvar}, left plot. The attributes are not too similar, since $\Delta \overline{G} > \tau$, but their difference is not big enough to make a split, since $\Delta \overline{G} < \epsilon$. The solution is to wait for more examples until $\epsilon$ (green triangle) decreases and is smaller than $\Delta \overline{G}$ (black star). Following this reasoning, $nmin= \ceil[\bigg]{\frac{R^2 \cdot ln(1/ \delta)}{2 \cdot (\Delta G)^2}}$, obtained by setting $\epsilon = \Delta \overline{G}$ in \eqref{eq:hoeff}, to guarantee that $\Delta \overline{G} \geq \epsilon$ will be satisfied in the next iteration, creating a split.
\end{theorem}
\vspace{0.4cm}
\begin{theorem}[$\Delta \overline{G} < \epsilon$ and $\Delta \overline{G} < \tau$ but $\epsilon > \tau$]
\label{scenar2o}
	The top attributes are very similar in terms of information gain, but $\epsilon$ is still higher than $\tau$, as can be seen in Figure~\ref{fig:icdm:epsilonvar}, right plot. The algorithm needs more instances so that $\epsilon$ (green triangle) decreases and is smaller than $\tau$ (red dot). Following this reasoning,  $nmin= \ceil[\bigg]{\frac{R^2 \cdot ln(1/ \delta)}{2 \cdot \tau^2}}$, by setting $\epsilon=\tau$ in \eqref{eq:hoeff} . In the next iteration $\epsilon \leq \tau$ will be satisfied, forcing a split. 
\end{theorem}

	The pseudocode of VFDT-\textit{nmin} is presented in Alg.~\ref{alg:vfdt-nmin}. The specific part of \textit{nmin adaptation} is shown in lines 22-26, where we specify how $nmin$ is going to be adapted based on the scenarios explained above. The idea is that, when those scenarios occur, we adapt the value of $nmin$, so that they don't occur in the next iteration, thus ensuring a split.
  In relation to the computational complexity of the \textit{nmin adaptation}, we can observe that this method does not add any overhead. Thus, the computational complexity of VFDT with \textit{nmin adaptation} is $O(n\cdot m)$.


\begin{algorithm}[H]

\caption{VFDT-\textit{nmin}: Very Fast Decision Tree with \textit{nmin adaptation}}\label{alg:vfdt-nmin}
\begin{algorithmic}[1]
\STATE $HT$: Tree with a single leaf (the root)
\STATE $X$: set of attributes
\STATE $G(\cdot)$: split evaluation function
\STATE $\tau$: hyperparameter set by the user
\STATE $nmin$: hyperparameter initially set by the user

\WHILE {stream is not empty}
  \STATE Read instance $I_i$
  \STATE Sort $I_i$ to corresponding leaf $l$ using $HT$
  \STATE Update statistics at leaf $l$
  \STATE Increment $n_l$: instances seen at $l$
  \IF {$nmin$ $\leq$ $n_l$}  
    \STATE Compute $\overline{G_l}(X_i)$ for each attribute $X_i$
    \STATE $X_a$, $X_b$ = attributes with the highest $\overline{G_l}$

    \STATE $\Delta \overline{G} = \overline{G}_l(X_a) -\overline{G}_l(X_b)$ 
    \STATE Compute $\epsilon$
    \IF {($\Delta \overline{G} > \epsilon$) or ($\epsilon < \tau$) }
      \STATE Replace $l$ with a node that splits on $X_a$ 
      \FOR {each branch of the split} 
        \STATE New leaf $l_m$ with initialized statistics
      \ENDFOR
    \ELSE
      \STATE Disable attr $\{X_{p} | (\overline{G}_l(X_p) -\overline{G}_l(X_a)) > \epsilon \}$ 
      \IF {$\Delta \overline{G} \leq \tau$}
          \STATE
          \STATE $nmin = \ceil[\bigg]{\frac{R^2 \cdot ln(1/ \delta)}{2 \cdot \tau^2}}$
      \ELSE
          \STATE $nmin = \ceil[\bigg]{\frac{R^2 \cdot ln(1/ \delta)}{2 \cdot (\Delta G)^2}}$
      \ENDIF
    \ENDIF
  \ENDIF
\ENDWHILE
\end{algorithmic}
\end{algorithm}




  \begin{figure}[!htb]
  \centering
  \includegraphics[width=0.49\textwidth]{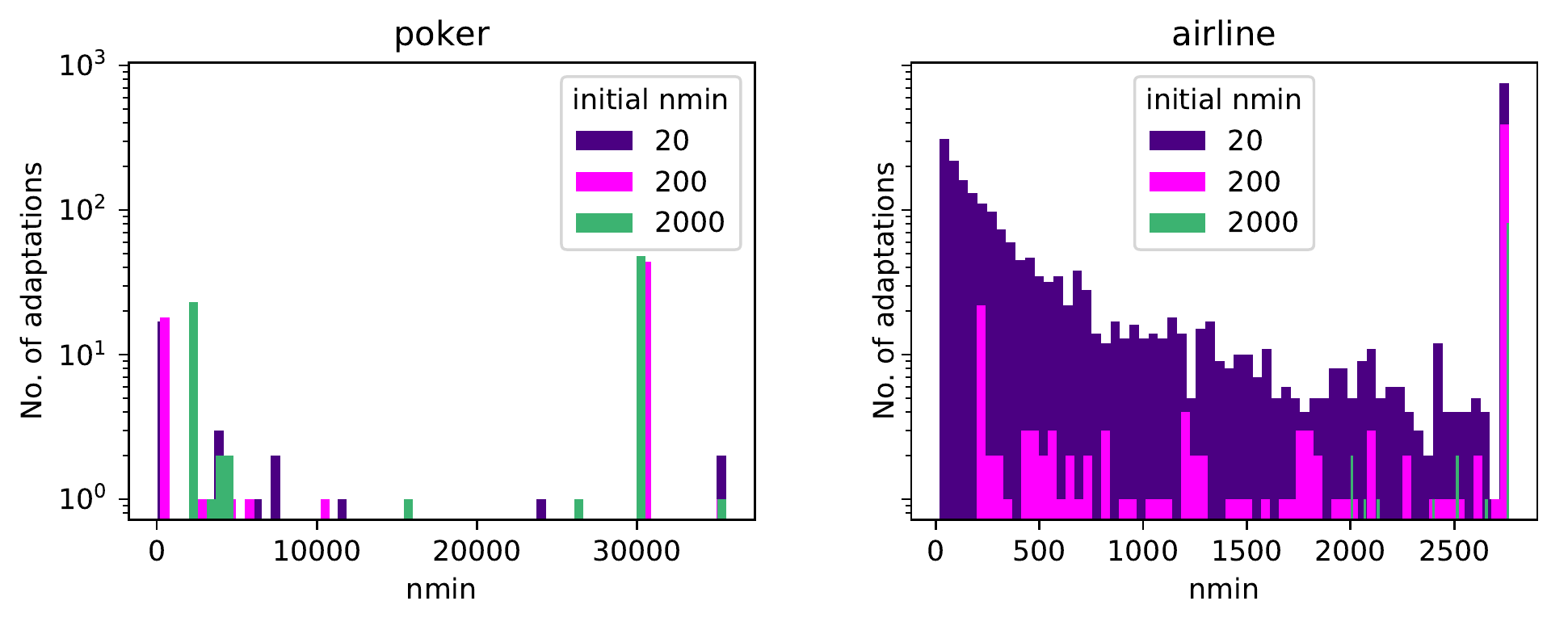}
  \caption{\small Variation of $nmin$ for $nmin$ initially set to $20,200,2000$ on poker and airline datasets (Table~\ref{tab:datasets}). 
  With a lower $nmin$, \textit{nmin adaptation} adapts $nmin$ to a higher value more frequently. The peaks on $nmin=2,763$ and $nmin=30,491$ is explained by Scenario~\ref{scenar2o}, since $\tau$ is a fixed hyperparameter.}
  \label{fig:nminvar}
  \end{figure}

  Finally, we show an example of how \textit{nmin adaptation} works for two of the datasets used in the final experiments. These datasets are described in Table~\ref{tab:datasets}.
  Figure~\ref{fig:nminvar} shows the $nmin$ variation for the cases when $nmin$ is initially set to $20,200,2000$. So, after those instances, $nmin$ will adapt to a higher value depending on the data observed so far at that specific leaf. The airline dataset shows many adaptations of $nmin$ when $nmin$ is initially set to $20$. This is expected, since we are showing the adaptations per leaf, so at the beginning all the leaves starting with $nmin=20$ will adapt that value to a much larger one. The same reasoning occurs when $nmin=200$ initially, since there will be less adaptations because the leaves need to wait for more instances, and there is a higher chance to split when more instances are read. 
	The poker dataset exhibits a different behavior, where $nmin$ adapts to a higher value, $30491$. This occurs in Scenario~\ref{scenar2o}, but since the poker dataset has 10 classes, the range $R$ of the Hoeffding bound equation \eqref{eq:hoeff} is higher. 
  Finally, looking at the cases where $nmin=2000$ (green), we observe how there is almost no adaptation. VFDT-\textit{nmin} either splits after $2000$ instances, or it adapts $nmin=2,763$ or $nmin=30,491$, because the attributes are very similar. 

\section{Energy Consumption of the VFDT}
\label{sec:energy_cons_vfdt}
Energy consumption is a necessary measurement for today's computations, since it has a direct impact on the electricity bill of data centers, and battery life of embedded devices. 
However, measuring energy consumption is a challenging task. As has been shown by researchers in computer architecture, estimating the energy consumption of a program is not straightforward, and is not as simple as measuring the execution time, since there are many other variables involved~\cite{Mazouz:2017dc}. 

In this section we first give a general background on energy consumption and its relationship to software energy consumption. We end the section with a more detailed view on the energy consumption in particular for the VFDT algorithm, presenting a theoretical energy model based on the number of instances of the stream, and number of numerical and nominal attributes. 

\subsection{General energy consumption}

Energy efficiency in computing usually refers to a hardware approach to reduce the power consumption of processors, or ways to make processors handle more operations using the same amount of power~\cite{koomey2009assessing}.

Power is the rate at which energy is being consumed. The average power during a time interval T is defined as~\cite{weste2005cmos}: 
  \begin{align}
      P_{avg} = \frac{E}{T}
  \end{align}
  where E, energy, is measured in joules (J), $P_{avg}$ is measured in watts (W), and time T is measured in seconds (s). 
  We can distinguish between dynamic and static power. Static power, also known as leakage power, is the power consumed when there is no circuit activity. Dynamic power, on the other hand, is the power dissipated by the circuit, from charging and discharging the capacitor~\cite{dubois2012parallel}:
  \begin{align}
  \label{eq:dyn_power}
        P_{dynamic} = \alpha \cdot C\cdot V_{dd}^2\cdot f
    \end{align}
    where $\alpha$ is the activity factor, representing the percentage of the circuit that is active. $V_{dd}$ is the voltage, $C$ the capacitance, and $f$ the clock frequency measured in hertz (Hz).
  Energy is the effort to perform a task, and it is defined as the integral of power over a period of time~\cite{dubois2012parallel}: 
  \begin{align}
  \label{eq:energy_time_power}
         E = \int_{0}^{T} P(t)dt 
  \end{align}
  In this study we focus on the measurement of energy consumption, since it gives an overview of how much power is consumed in an interval of time.

  Finally, we conclude with an explanation of how programs consume energy. The total execution time of a program is defined as~\cite{dubois2012parallel}:
  \begin{align}
      \label{eq:time}
      T_{exe} = IC \times CPI \times T_{c}
    \end{align}
    where IC is the number of executed instructions, CPI (clock cycles per instruction) is the average number of clock cycles needed to execute each instruction, and $T_C$ is the clock cycle time of the processor.
    The total energy consumed by a program is:
    \begin{align}
      \label{eq:eng_cpi}
      E = IC \times CPI \times EPC
    \end{align} 
    where EPC is the energy per clock cycle, and it is defined as
    \begin{align}
      \label{eq:eng_int}
      EPC \propto C \cdot V_{dd}^2
    \end{align}
The value CPI depends on the type of instruction, since different instructions require different number of clock cycles to complete. However, measuring only time does not give a realistic view on the energy consumption, because there are instructions that can consume more energy due to a long delay (e.g. memory accesses), or others that consume more energy because of a high requirement of computations (floating point operations). Both could obtain similar energy consumption levels, however, the first one would have a longer execution time than the last one.

\subsection{VFDT energy model}
\label{sec:vfdt_energymodel}
The energy model of the VFDT is based on the energy consumption of the main functions of the algorithm. 
These functions are taken from the pseudocode of the VFDT~\cite{domingos2000mining}. Alg.~\ref{alg:vfdt-nmin} shows the pseudocode for the VFDT algorithm with the \textit{nmin adaptation} functionality added, but the main functions can also be observed there. 
The main functions are the following:
\begin{itemize}
	\item \textbf{Sort instance to leaf}. When an instance is read, the first step is to traverse the tree based on the attribute values of that instance, to reach the correspondent leaf. 
	\item \textbf{Update attributes}: Once the leaf is reached, the information at that leaf is updated with the attribute/class information of the instance. The update process is different if the attribute is numerical or nominal. For nominal attributes a simple table with the counts is needed. For updating the numerical attribute the mean and the standard deviation are updated.
	\item \textbf{Update instance count}: After each instance is read the counter at that leaf is updated.
 
	\item \textbf{Calculate entropy}: Once \textit{nmin} instances are observed at a leaf, the entropy (information gain in this case) is calculated for each attribute. 
	\item \textbf{Get best attribute}: The attributes with the highest information gain are chosen. 
	\item \textbf{Calculate Hoeffding Bound}: We then compare the difference between the best and the second best attribute with the Hoeffding bound, calculated with \eqref{eq:hoeff}.
	\item \textbf{Create new node}: If there is a clear attribute to split on, we split on the best attribute creating a new node. 
\end{itemize}

 Based on the information provided above, we present the energy consumption of the VFDT in the following model:
 \begin{equation} \label{eq:eng_vfdt}
	E_{VFDT}= E_{comp} + E_{cache\_tot} + E_{cache\_miss\_tot},
\end{equation}
where  $E_{comp}$ is the energy consumed on computations, $E_{cache\_tot}$ is the energy consumed on cache accesses, and $E_{cache\_miss\_tot}$ is the energy consumed on cache misses. 
They are defined as follows:
\begin{equation} \label{eq:eng_comp}
	E_{comp} = n_{FPU}\cdot E_{FPU} + n_{INT}\cdot E_{INT}, 
\end{equation}
where $n_{FPU}$ is the number of floating point operations, $E_{FPU}$ is the average energy per floating point operation, $n_{INT}$ is the number of integer operations, and $E_{INT}$ is the average energy per integer operation. 


\begin{equation} \label{eq:eng_cache}
	E_{cache\_tot} = n_{cache}\cdot E_{cache},
\end{equation}
where $n_{cache}$ is the number of accesses to cache, and $E_{cache}$ is the average energy per access to cache. 
Finally, 
\begin{equation} \label{eq:eng_cache_miss}
	E_{cache\_miss\_tot} = n_{cache\_miss}\cdot (E_{cache\_miss} + E_{DRAM}),
\end{equation}
where $n_{cache\_miss}$ is the number of cache misses, $E_{cache\_miss}$ is the average energy per cache miss, and $E_{DRAM}$ is the average energy per DRAM access.


The next step is to map these $n_{FPU}$, $n_{INT}$, $n_{cache}$, and $n_{cache\_miss}$ to the VFDT algorithm's functions, explained at the beginning of this section. 

\begin{IEEEeqnarray}{rCl}
\label{eq:nfpu}
n_{FPU}&=&n_{comp}(updating\_numerical\_atts) \IEEEnonumber \\
		&&+\:n_{comp}(calc\_entropy) \IEEEnonumber \\
 		&&+\:n_{comp}(calc\_hoeff\_bound)  \\
 		&&+\:n_{comp}(get\_best\_att) \IEEEnonumber
\end{IEEEeqnarray}
\begin{IEEEeqnarray}{rCl}
\label{eq:nint}
n_{INT}&=&n_{comp}(updating\_nominal\_atts) \IEEEnonumber \\
		&&+\:n_{comp}(updating\_instance\_count),  
\end{IEEEeqnarray}
where $n_{comp}(f_i)$ refers to the number of computations required by function $f_i$.


\begin{IEEEeqnarray}{rCl}
\label{eq:n_cache}
n_{cache}&=n_{acc}(updating\_atts)  
\end{IEEEeqnarray}

\begin{IEEEeqnarray}{rCl}
\label{eq:n_cache_miss}
n_{cache\_miss}&=&n_{acc}(sorting\_instance\_to\_leaf) \IEEEnonumber \\
		&&+\:n_{acc}(updating\_atts) \IEEEnonumber \\
		&&+\:n_{acc}(calc\_entropy) \IEEEnonumber \\
		&&+\:n_{acc}(calc\_hoeff\_bound)  \\
		&&+\:n_{acc}(new\_node), \IEEEnonumber 
\end{IEEEeqnarray}
where $n_{acc}(f_i)$ represents the number of accesses to memory or cache in order to execute function $f_i$. 
The number of cache and memory accesses of updating the attributes ($updating\_atts$) will depend on the block size of the cache. If the block size is big enough, then we would have one cache miss to update the information of the first attribute, and then cache hits for the rest of the attributes. However, if there are many attributes, thus not fitting on the block size $B$, then there will be a cache miss for every attribute that exceeds the block size.
We also consider the presence of a cache miss every time a node of the tree is traversed, and every time we calculate the entropy and Hoeffding bound values. 

The last step is to express these number of accesses and computations based on the number of instances ($N$), the $nmin$ value, the number of numerical attributes ($A_{f}$), the number of nominal attributes ($A_{i}$), and the block cache size $B$. We then obtain the following:

\begin{IEEEeqnarray}{rCl}
\label{eq:nfpu2}
n_{FPU}&=&N\cdot A_{f} + \frac{N}{nmin} \cdot (A_{f} + A_{i}) \IEEEnonumber\\
	   &&+\: \frac{N}{nmin} + \frac{N}{nmin} \cdot (A_{f} + A_{i}) \\
       &=&\:N\cdot A_{f} + 2\cdot \frac{N}{nmin} \cdot (A_{f} + A_{i}) + \frac{N}{nmin} \IEEEnonumber
\end{IEEEeqnarray}
Updating numerical attributes is one access per instance per numerical attribute; calculating the entropy is one access per attribute (thus the sum of nominal and numerical attributes) every $nmin$ instances; calculating the Hoeffding bound is one access every $nmin$ instances; and calculating the best attribute is the same as calculating the entropy.

\begin{IEEEeqnarray}{rCl}
\label{eq:nint2}
n_{INT}&=&N\cdot A_{i} + N  
\end{IEEEeqnarray}
Updating nominal attributes is, as before, one access per instance per nominal attribute; and one access per instance for updating the counter.


\begin{IEEEeqnarray}{rCl}
\label{eq:ncache2}
n_{cache}&=&N\cdot (A_{f} + A_{i} - \frac{A_{f} + A_{i}}{B}) 
\end{IEEEeqnarray}
To update the attributes, we consider one cache hit per all attributes per instance, minus all the attributes that don't fit on the block size $B$ and create cache misses.

\begin{IEEEeqnarray}{rCl}
\label{eq:n_cache_miss2}
n_{cache\_miss}&=&N\cdot (A_{f} + A_{i} + \frac{A_{f} + A_{i}}{B})  \IEEEnonumber \\
		&&+\: \frac{N}{nmin} + \frac{N}{nmin} + \frac{N}{nmin}  \\
       &=&\:N\cdot (A_{f} + A_{i} +  \frac{A_{f} + A_{i}}{B}) + 3\cdot \frac{N}{nmin} \IEEEnonumber
\end{IEEEeqnarray}
To calculate the number of accesses of sorting an instance to a leaf we assume that we need to access one level per attribute, which is the worst case scenario. So the total number of accesses in this case is one per instance per attribute. 
To update the attributes, as was explained before, it's one miss per all attributes that exceed the block size $B$, per instance.
Finally, to access the needed values to calculate the entropy, the Hoeffding bound, and to split, we consider one access every $nmin$ instances.

Based on \eqref{eq:eng_vfdt}, \eqref{eq:eng_comp}, \eqref{eq:eng_cache}, \eqref{eq:eng_cache_miss}, \eqref{eq:nfpu2}, \eqref{eq:nint2}, \eqref{eq:ncache2}, and \eqref{eq:n_cache_miss2}, our final energy model equation is the following:
\begin{IEEEeqnarray}{rCl}
\label{eq:VFDT_finalmodel}
E_{VFDT}&=& E_{FPU}\cdot(N\cdot A_{f} + 2\cdot \frac{N}{nmin} \cdot (A_{f} + A_{i}) \IEEEnonumber \\
	   &&+\: \frac{N}{nmin}) +  E_{INT}\cdot( N\cdot A_{i} + N ) \IEEEnonumber \\
	   &&+\: E_{cache}\cdot(N\cdot (A_{f} + A_{i} - \frac{A_{f} + A_{i}}{B} )) \IEEEnonumber \\
	   &&+\: (E_{cache\_miss} + E_{DRAM} )\cdot(N\cdot(A_{f} + A_{i} \IEEEnonumber \\ 
	   &&+\: \frac{A_{f} + A_{i}}{B}) + 3\cdot \frac{N}{nmin})    
\end{IEEEeqnarray}

This is a general and simplified model of how the VFDT algorithm consumes energy. The energy values (i.e. $E_{cache}$, $E_{FPU}$, $E_{INT}$, $E_{DRAM}$, and $E_{cache\_miss}$) will vary depending on the processor and architecture, although there is a lot of research that ranks these operations based on their energy consumption~\cite{horowitz20141}. For instance, a DRAM instruction consumes three orders of magnitude more energy than an ALU operation. 
We can see the importance of the number of attributes in the overall energy consumption of the algorithm. Since $E_{FPU}$ is significantly higher than $E_{INT}$, numerical attributes have a higher impact on energy consumption than nominal attributes.

\section{Experimental Design}
  \label{sec:experimental_design}
  We have designed an experiment that compares VFDT, VFDT-\textit{nmin}, and CVFDT (Concept-Adapting Very Fast Decision Tree~\cite{hulten2001mining}). The goal of this experiment is to compare the energy consumption and accuracy of all algorithms. Since CVFDT is designed for concept drift scenarios, we also analyze the possible trade-off between accuracy and energy consumption. Namely, how much more energy is CVFDT consuming to be able to achieve a higher accuracy in concept drift scenarios. 
  We have a set of concept drift datasets to test this phenomenon. 

  We run the experiments on a machine with an 3.5 GHz Intel Core i7, with 16GB of RAM, running OSX. 
  To estimate the energy consumption we use Intel Power Gadget\footnote{\url{https://software.intel.com/en-us/articles/intel-power-gadget-20}}, that accesses the performance counters of the processor, together with Intel's RAPL interface to obtain energy consumption estimations. 
  The implementation of VFDT-\textit{nmin} together with the scripts to conduct the experiments are available online\footnote{\url{https://github.com/egarciamartin/hoeffding-nmin-adaptation}}.

  \subsection{Datasets}
  \label{sec:datasets}
  We used real and artificial datasets, inspired by the work from~\cite{bifet2017extremely}.
  The datasets are described in Table~\ref{tab:datasets}. There are a total of 15 datasets, 12 artificial datasets generated with Massive Online Analysis (MOA)~\cite{DBLP:journals/jmlr/BifetHKP10}, and 3 real world datasets. The artificial datasets are the following:


  \begin{table}[htb]
  \renewcommand{\arraystretch}{1.2}
  \setlength{\tabcolsep}{4.5pt}

  \centering
  \scriptsize
  \caption{\small Datasets used in the experiment to compare VFDT, VFDT-\textit{nmin}, and CVFDT. $A_{i}$  and $A_{f}$ represent the number of nominal and numerical attributes, respectively. The details of each dataset is presented in Section~\ref{sec:datasets}}
  \label{tab:datasets}

  \begin{tabular}{l r r r r r }

  \toprule

   Dataset      & Train & Test    & $A_{i}$  & $A_{f}$ & Class  \\
   \midrule
    HYP(0.0001) & 670,000 & 330,000   &  0     & 10    & 5  \\
     HYP(0.001) & 670,000 & 330,000   &  0     & 10    & 5  \\
         LED(1) & 670,000 & 330,000   &  24    & 0     & 10 \\
         LED(2) & 670,000 & 330,000   &  24    & 0     & 10 \\
      RBF(10,0) & 670,000 & 330,000   &  0     & 10    & 5  \\ 
 RBF(10,0.0001) & 670,000 & 330,000   &  0     & 10    & 5  \\ 
  RBF(10,0.001) & 670,000 & 330,000   &  0     & 10    & 5  \\ 
      RBF(50,0) & 670,000 & 330,000   &  0     & 10    & 5  \\
 RBF(50,0.0001) & 670,000 & 330,000   &  0     & 10    & 5  \\
  RBF(50,0.001) & 670,000 & 330,000   &  0     & 10    & 5  \\
        SEA(10) & 670,000 & 330,000   &  0     & 3     & 2  \\
        SEA(20) & 670,000 & 330,000   &  0     & 3     & 2  \\ 
        airline & 361,387 & 177,996   & 4      & 3     & 2  \\
    electricity & 30,359  & 14,953    & 1      & 6     & 2  \\
          poker & 555,564 & 273,637   & 5      & 5     & 10 \\
   \bottomrule
  \end{tabular}
  \end{table}


\paragraph*{HYP($v$)}
Hyperplane dataset.
This dataset is generated by creating a set of points that satisfy 
  $\sum_{i=1}^{d}w_ix_i = w_0$
 ,where $x_i$ is the coordinate for each point. 
 Then, examples that satisfy $\sum_{i=1}^{d}w_ix_i \geq x_0$ are labeled as positive, and examples that satisfy $\sum_{i=1}^{d}w_ix_i < x_0$ are labeled as negative.
 Drift is introduced to each weight ($w_i$), and the amount of change is represented by $v$.
 More details are given in~\cite{hulten2001mining}.

 \paragraph*{LED($x$)}
LED dataset with $x$ attributes with drift. The goal is to predict the digit on a LED display with seven segments, where each attribute has a 10\% chance of being inverted~\cite{breiman2017classification}. 

 \paragraph*{RBF($x,v$)}
  The radial based function (RBF) artificial dataset has 10 numerical attributes. The generator creates $x$ number of centroids, each with a random center, class label and weight. Each new example randomly selects a center, considering that centers with higher weights are more likely to be chosen. The chosen centroid represents the class of the example. Drift is introduced by moving the centroids with speed $v$. More details are given by~\cite{bifet2017extremely}.

 \paragraph*{SEA($v$)}
 The SEA artificial dataset was first introduced by~\cite{street2001streaming} to test abrupt concept drift. 
 The $v$ value represents the percentage of noise introduced. It has 3 numerical attributes with a range between 0 and 10. For each example, the first two attributes are summed and compared against a threshold value ($\theta$).


The explanations above have been based on the work by~\cite{bifet2017extremely}, where they use a similar set of datasets to compare different machine learning frameworks. 

We also tested three real datasets, all available from the MOA official website~\cite{moaDatasets}.
The poker dataset is a normalized dataset available from the UCI repository. Each instance represents a hand consisting of five playing cards, where each card has two attributes; suit and rank. 

The electricity dataset is originally described in~\cite{harries1999splice}, and is frequently used in the study of performance comparisons. Each instance represents the change of the electricity price based on different attributes such as day of the week, represented by the Australian New South Wales Electricity Market.

Finally, the airline dataset is provided by Elena Ikonomovska~\cite{elenaData} and the task is to predict if a given flight will be delayed based on attributes such as airport of origin and airline. 

\subsection{Algorithms and setups}

We compare VFDT, VFDT-\textit{nmin}, and CVFDT under the mentioned datasets. The initial value of $nmin$ has been set to 200, which was the default value used by the original authors.
We evaluate all algorithms based on the following measures: accuracy (\% of correctly classified instances), energy consumed by the processor, and energy consumed by the DRAM. We evaluate the accuracy by having a training set and a test set that is different from the training set, as can be observed in Table~\ref{tab:datasets}. We have not performed yet prequential evaluation as with this method, however that is planned for future works.

\section{Results}
\label{sec:results}
The results of the experiments are shown in Table~\ref{tab:res}. 
These results are obtained from running the algorithms VFDT, VFDT-\textit{nmin}, and CVFDT under the datasets shown in Table~\ref{tab:datasets}. We have evaluated the accuracy (percentage of correctly classified instances) and the energy consumption of 10 runs, and averaged the results.
We have measured the total energy consumption as the sum of the energy consumed by the processor and the energy consumed by the DRAM, since that is the output given by the tool.

\begin{table*}[htb]
  \centering
  \renewcommand*{\arraystretch}{1.3}
  \scriptsize
  \caption{Energy consumption and accuracy results. Algorithms: VFDT, VFDT-\textit{nmin}, and CVFDT. Measurements: Accuracy, Total energy, processor energy, dram energy. Total energy = processor energy + dram energy. Higher accuracy and lower total energy consumption values for each dataset are presented in bold.}
  \label{tab:res} 
  \input{results_tex.tex}
\end{table*}

 In order to have a better understanding of the results, we have created Table~\ref{tab:diff_VFDT}, where we compute the difference in accuracy and energy between VFDT and VFDT-\textit{nmin}, and between VFDT-\textit{nmin} and CVFDT. The difference in accuracy is measured by subtracting the accuracy of VFDT-\textit{nmin}, minus the accuracy of VFDT (or CVFDT depending on the column). Thus, a positive value in such column shows that VFDT-\textit{nmin} obtained a higher accuracy than the compared algorithm. 
 The difference in energy represents the percentage of energy reduced between VFDT-\textit{nmin} and the compared algorithm. A negative value represents that we reduced the energy by that percentage. For instance, VFDT-\textit{nmin} consumed 20.49\% less energy than VFDT in the HYP(0.0001) dataset. 

\begin{table}[htb]
  \centering
  \renewcommand*{\arraystretch}{1.3}
  \scriptsize
  \caption{Difference in accuracy ($\Delta$Acc) and energy consumption ($\Delta$Energy) between VFDT and VFDT-nmin; and VFDT-nmin and CVFDT. A positive number in accuracy means that VFDT-nmin obtained a higher accuracy. A negative number in energy means that the VFDT-nmin reduced the energy consumption by that percentage. Higher accuracy and lower energy consumption of the VFDT-\textit{nmin} are presented in bold}
  \label{tab:diff_VFDT} 
  \input{table_diff.tex}
\end{table}

\section{Discussion}
\label{sec:discussion}
The discussion of the results is focused, first, on the energy comparison between the CVFDT and VFDT to VFDT-nmin. Then, we analyze the difference in accuracy between the mentioned algorithms. We conclude the discussions with an analysis of the impact of the number of numerical attributes in the overall energy consumption, linking the results to the energy model proposed in Section~\ref{sec:vfdt_energymodel}.

\begin{figure}[htb]
\centering
\includegraphics[width=0.48\textwidth]{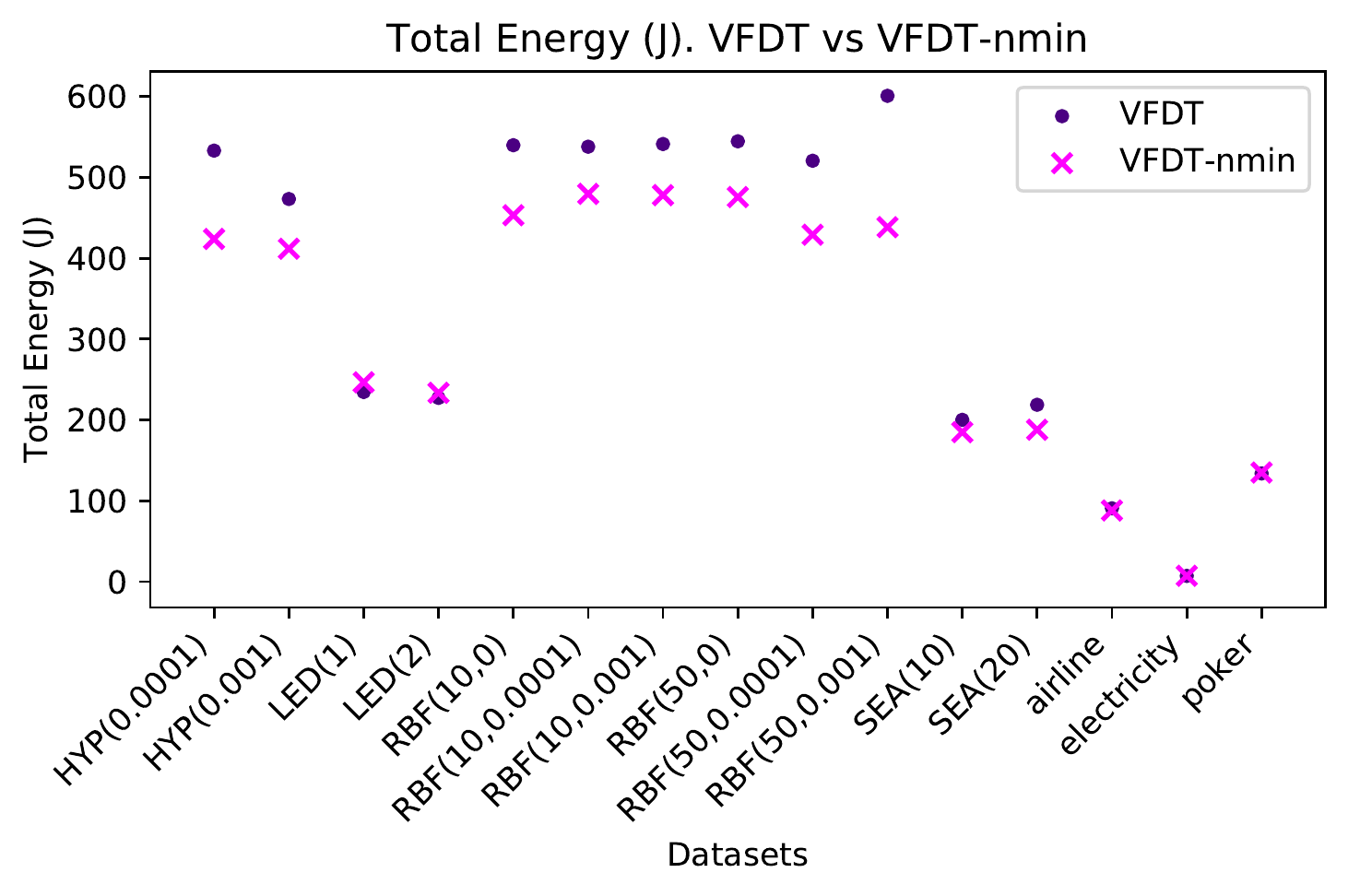}
\caption{\small VFDT and VFDT-nmin total energy comparison. We observe that VFDT-nmin obtains a lower energy consumption in 11 out of 15 datasets. }
\label{fig:datasets_total_eng}
\end{figure}

\begin{figure}[htb]
\centering
\includegraphics[width=0.48\textwidth]{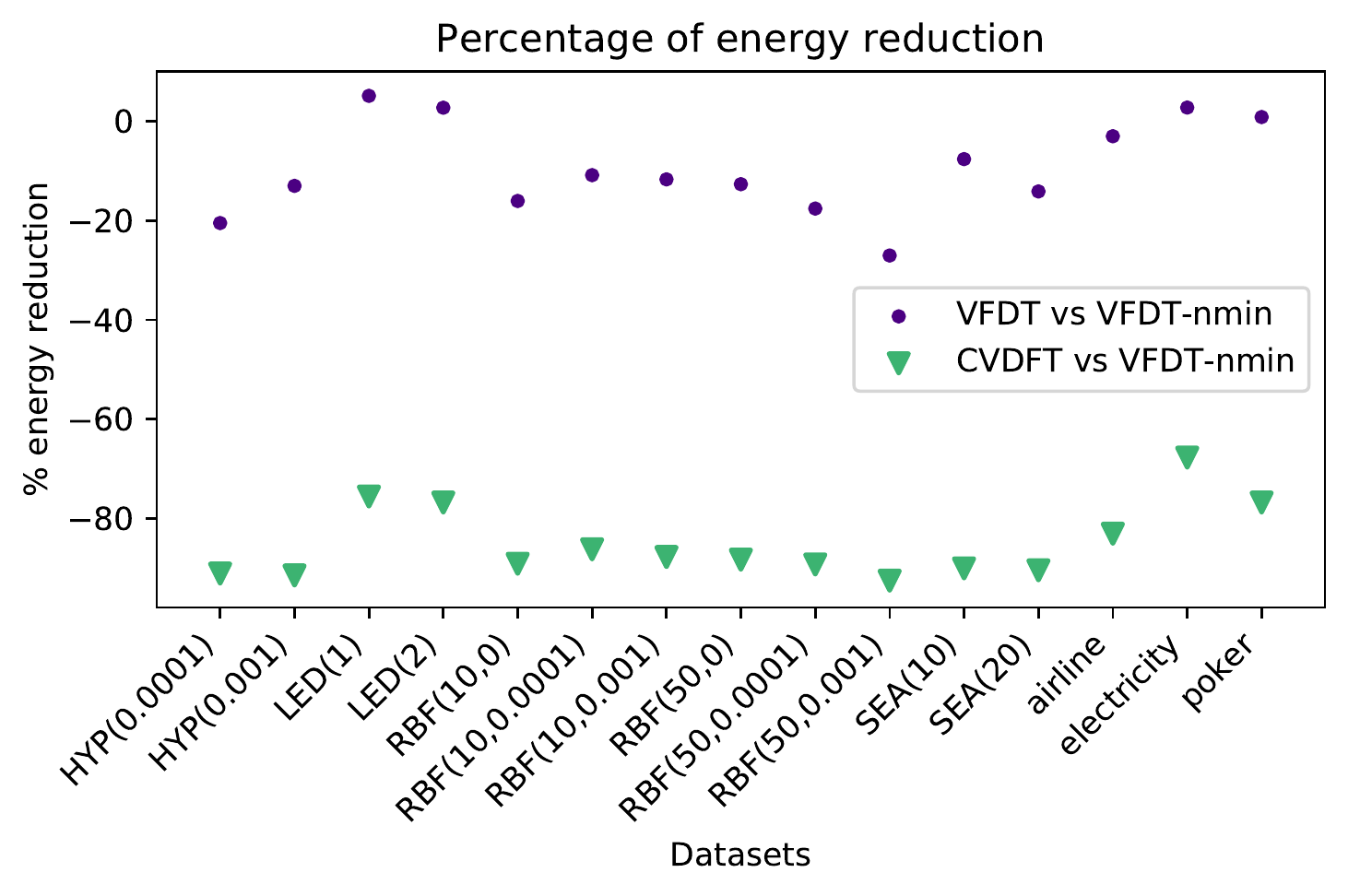}
\caption{\small VFDT vs VFDT-nmin percentage of reduced energy. Lower is better, since it means a higher energy reduction from VFDT-\textit{nmin}. For instance, VFDT-nmin reduced the energy consumption by 20\% for the HYP(0.0001) dataset. We observe how VFDT-nmin reduces the energy consumption by a high percentage in comparison to the CVFDT algorithm.}
\label{fig:datasets_perc_eng}
\end{figure}

 The results show that VFDT-\textit{nmin} consumes significantly less energy than VFDT in most of the datasets (11/15), with a maximum energy reduction of 27\% of energy (RBF(50,0.0001) dataset). If we compare VFDT-\textit{nmin} to CVFDT, this difference is considerably larger. On average, VFDT-\textit{nmin} consumes 85\% less energy than CVDFT. This is visible in Figures~\ref{fig:datasets_total_eng} and \ref{fig:datasets_perc_eng}. Figure~\ref{fig:datasets_total_eng} shows the energy consumption of VFDT and VFDT-\textit{nmin} for all datasets. We can observe how VFDT-\textit{nmin} either obtains a lower energy consumption than VFDT, or a very similar value. 
 Figure~\ref{fig:datasets_perc_eng} shows the comparison on percentage of energy reduction between the three algorithms. This last comparison portrays the large energy savings from VFDT-\textit{nmin} compared to CVFDT. 	
 We also observe that VFDT-\textit{nmin} obtains higher energy consumption than VFDT in two of the three real world datasets (electricity and poker). Although this difference in energy consumption is minimal (2.78\% in the electricity dataset and 0.87\% in the poker dataset), we plan to investigate this further with more real world datasets.

\begin{figure}[htb]
\centering
\includegraphics[width=0.48\textwidth]{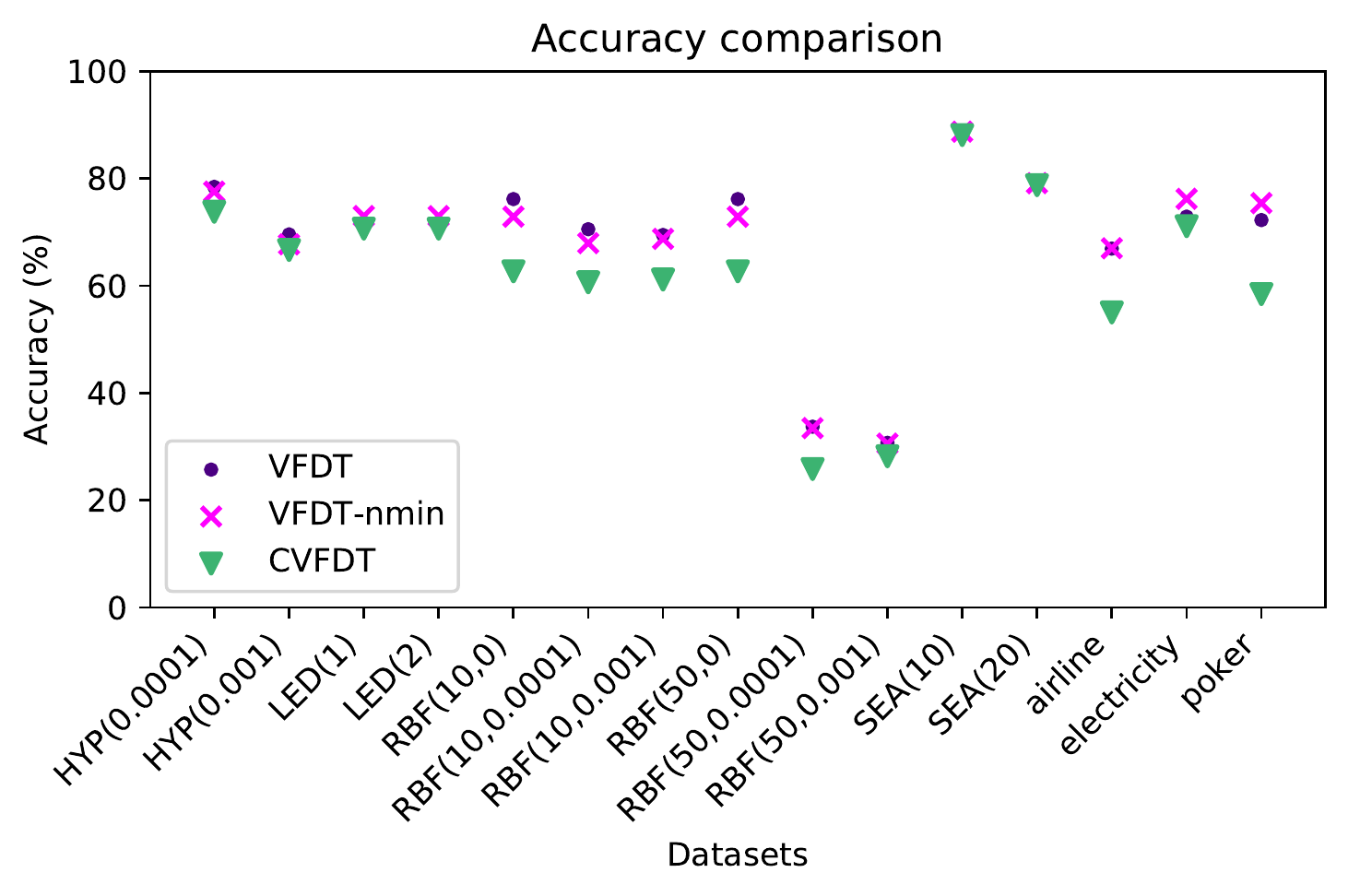}
\caption{\small Comparison in accuracy between VFDT, VFDT-nmin, and CVFDT. VFDT and VFDT-\textit{nmin} obtain very similar levels of accuracy. CVFDT obtain significantly lower accuracy values. }
\label{fig:datasets_acc}
\end{figure}

The next variable to analyze is accuracy. We would expect CVFDT to obtain higher accuracy at the expense of the higher energy consumption, since CVFDT is meant to perform better in concept drift datasets.
However, the results show that CVFDT obtained lower accuracy compared to the other two algorithms, even for datasets with concept drift.
In all cases, VFDT and VFDT-\textit{nmin} obtained higher values of accuracy. 
Figure~\ref{fig:datasets_acc} shows the accuracy comparison between VFDT, VFDT-\textit{nmin}, and CVFDT. We observe that the accuracy of VFDT and VFDT-\textit{nmin} is very similar, VFDT obtaining 0.22\% higher accuracy on average.


\begin{figure}[htb]
\centering
\includegraphics[width=0.48\textwidth]{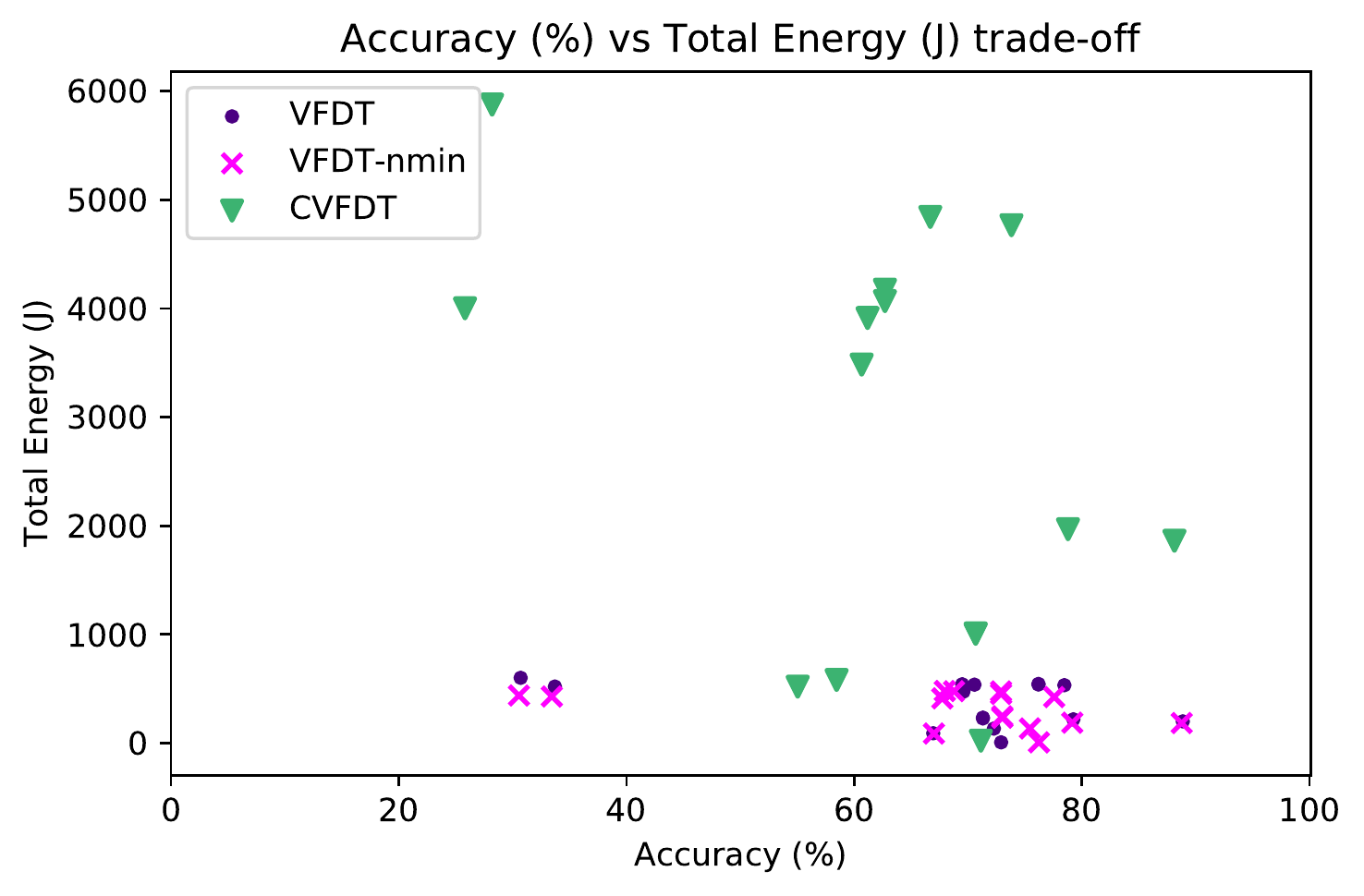}
\caption{\small Relationship between accuracy and energy consumption, for all datasets and all algorithms
The optimal scenario is at the right bottom, with a low energy consumption and high accuracy. We can observe how CVFDT consumes significantly more energy in comparison to VFDT and VFDT-nmin, without increasing accuracy. }
\label{fig:acc_eng_tradeoff}
\end{figure}

Figure~\ref{fig:acc_eng_tradeoff} shows the relationship between accuracy and energy consumption. The optimal data points lie at the bottom right of the figure, representing low energy consumption and high accuracy. Almost all VFDT-\textit{nmin} executions lie in the low energy consumption / high accuracy range. However, we can observe how the points representing the CVFDT executions are predominant towards high energy consumption and low accuracy areas (top left). 
Although the figure shows no apparent trade-off between accuracy and energy consumption, the results in Table~\ref{tab:res} show that those datasets where VFDT-\textit{nmin} obtained a higher accuracy (LED, electricity, and poker), it obtained also lower energy consumption. This suggests a trade-off between accuracy and energy consumption, where in order to achieve a higher accuracy, more energy needs to be spent.

Finally, regarding numerical attributes, we can observe that the number of numerical attributes directly affects the energy consumption significantly more than nominal attributes (energy model in Section~\ref{sec:vfdt_energymodel}). The reason is that the average energy per floating point operation ($E_{FPU}$) is significantly higher than the average energy per integer instruction ($E_{INT}$)~\cite{horowitz20141}. 
Moreover, storing the statistics of floating values (numerical values) takes up more space than storing the statistics of integer values (nominal attributes).
If we take a look at Table~\ref{tab:res}, at datasets LED and HYP (independent of the particular parameters), we can observe how, with the same number of instances, LED consumes approximately half of the energy of HYP. LED has 24 nominal attributes (Table~\ref{tab:datasets}) and HYP has 10 numerical attributes. The interesting phenomena is that even though LED has more than double the number of attributes, HYP still consumes double the energy because of the high energy consumption impact of having numerical attributes. 
This result opens a new direction for future works, to implement a more energy efficient approach to handle numerical attributes for streaming scenarios. 

In summary, VFDT-\textit{nmin} consumes 9.50\% less energy than VFDT, while only sacrificing less than 1\% of accuracy. The highest energy reduction occurs for the dataset RBF(50,0.001), where VFDT-\textit{nmin} consumes 27\% less energy than VFDT, sacrificing 0.15\% of accuracy. 
These results show that VFDT-\textit{nmin} is able to obtain competitive results in terms of accuracy, while being able to significantly reduce its energy consumption.

\section{Conclusions}
\label{sec:conclusions}

In this paper we introduced \textit{nmin adaptation} for Hoeffding trees to reduce their energy consumption.
We compared VFDT-\textit{nmin} (VFDT with \textit{nmin adaptation}) to the standard VFDT and CVFDT, under 15 datasets. The results showed that VDFT-\textit{nmin} consumes up to 27\% less energy, affecting accuracy at most by a $3\%$, in comparison with the standard VFDT. In comparison to CVFDT, VFDT-\textit{nmin} consumes 85\% less energy, obtaining 6\% higher accuracy values, on average.

We have shown a way to reduce the energy consumption of the VFDT, by first identifying the source of unnecessary computations with a theoretical energy model of the VFDT. Based on that information, we have reduced the amount of unnecessary computations, thus reducing the overall energy consumption,  while only marginally affecting accuracy.
We believe that this study presents a significant contribution to the field of data stream mining. We illustrate a method that can change the way we currently design this class of algorithms, with a new focus on energy efficiency and dynamic parameter adaptation. 
Algorithms with low energy consumption are necessary for embedded systems and other resource constrained devices; and desirable for platforms that require many computations, such as data centers.

For future work, we aim to evaluate further the \textit{nmin adaptation} method on other Hoeffding tree algorithms, such as the Hoeffding Adaptive Tree (\cite{bifet2009adaptive}). As was already mentioned, we plan to investigate more energy efficient ways to handle numerical attributes in streaming scenarios.

\bibliographystyle{IEEEtran}
\bibliography{lib}

\end{document}

%% file: results_tex.tex
\begin{tabular}{lrrrrrrrrrrrr}
\toprule
        Dataset & \multicolumn{3}{c}{Accuracy (\%)} & \multicolumn{3}{c}{Total Energy(J)} & \multicolumn{3}{c}{Proc Energy (J)} & \multicolumn{3}{c}{DRAM Energy(J)} \\
\cmidrule(lr){2-4} \cmidrule(lr){5-7} \cmidrule(lr){8-10} \cmidrule(lr){11-13}
 & VFDT-{nmin} & CVFDT &  VFDT &  VFDT-{nmin} & CVFDT &   VFDT & VFDT-{nmin} &   CVFDT &   VFDT & VFDT-{nmin} &  CVFDT &  VFDT \\ 
\midrule
    HYP(0.0001) &        77.57 & 73.81 & \textbf{78.46} &          \textbf{423.86} & 4766.54 & 533.11 &          407.58 & 4530.18 & 509.51 &          16.27 & 236.36 & 23.60 \\
     HYP(0.001) &        67.74 & 66.69 & \textbf{69.59} &          \textbf{411.77} & 4842.61 & 473.28 &          395.56 & 4619.29 & 454.57 &          16.21 & 223.32 & 18.72 \\
         LED(1) &        \textbf{73.01} & 70.67 & 71.31 &          246.65 & 1010.25 & \textbf{234.59} &          237.37 &  985.05 & 225.48 &           9.28 &  25.19 &  9.11 \\
         LED(2) &        \textbf{73.01} & 70.67 & 71.31 &          233.51 & 1006.40 & \textbf{227.26} &          223.98 &  981.64 & 218.06 &           9.53 &  24.76 &  9.19 \\
      RBF(10,0) &        72.90 & 62.71 & \textbf{76.19} &          \textbf{453.23} & 4173.32 & 539.79 &          436.59 & 3985.00 & 519.64 &          16.63 & 188.33 & 20.15 \\
 RBF(10,0.0001) &        67.96 & 60.66 & \textbf{70.56} &          \textbf{479.64} & 3483.51 & 537.97 &          462.53 & 3352.17 & 518.05 &          17.11 & 131.34 & 19.92 \\
  RBF(10,0.001) &        68.77 & 61.18 & \textbf{69.49} &          \textbf{477.99} & 3912.56 & 541.28 &          460.83 & 3746.63 & 520.60 &          17.16 & 165.92 & 20.68 \\
      RBF(50,0) &        72.90 & 62.71 & \textbf{76.19} &          \textbf{475.70} & 4068.16 & 544.64 &          458.65 & 3883.77 & 524.35 &          17.05 & 184.39 & 20.30 \\
 RBF(50,0.0001) &        33.45 & 25.82 & \textbf{33.71} &          \textbf{429.16} & 4001.94 & 520.61 &          412.47 & 3830.61 & 500.11 &          16.69 & 171.33 & 20.50 \\
  RBF(50,0.001) &        30.55 & 28.20 & \textbf{30.71} &          \textbf{438.46} & 5876.57 & 600.89 &          419.91 & 5576.47 & 575.06 &          18.55 & 300.10 & 25.83 \\
        SEA(10) &        88.77 & 88.13 & \textbf{88.87} &          \textbf{184.90} & 1862.83 & 200.14 &          177.74 & 1783.54 & 192.40 &           7.16 &  79.29 &  7.74 \\
        SEA(20) &        79.16 & 78.78 & \textbf{79.25} &          \textbf{187.88} & 1968.82 & 218.77 &          180.34 & 1891.36 & 209.48 &           7.53 &  77.46 &  9.29 \\
        airline &        \textbf{67.01} & 55.04 & 66.94 &           \textbf{88.01 }&  520.11 &  90.73 &           85.21 &  498.03 &  87.85 &           2.80 &  22.07 &  2.89 \\
    electricity &        \textbf{76.23} & 71.13 & 72.91 &            7.32 &   22.70 &   \textbf{7.12} &            7.13 &   21.95 &   6.91 &           0.19 &   0.75 &  0.21 \\
          poker &        \textbf{75.44} & 58.46 & 72.27 &          134.97 &  581.77 & \textbf{133.81} &          129.73 &  567.20 & 128.65 &           5.25 &  14.57 &  5.16 \\
\bottomrule
\end{tabular}

%% file: table_diff.tex
\begin{tabular}{lrrrr}
\toprule
& \multicolumn{2}{c}{VFDT-nmin vs VFDT }& \multicolumn{2}{c}{VFDT-nmin vs CVFDT} \\ 
\cmidrule(lr){2-3} \cmidrule(lr){4-5}
Dataset & $\Delta$Acc (\%) & $\Delta$Energy(\%) & $\Delta$Acc (\%) & $\Delta$Energy(\%) \\ 
\midrule
    HYP(0.0001) &         -0.90 & \textbf{-20.49} &  \textbf{3.75} & \textbf{-91.11} \\
     HYP(0.001) &         -1.85 & \textbf{-13.00} &  \textbf{1.05} & \textbf{-91.50} \\
         LED(1) &  \textbf{1.70}&           5.14  &  \textbf{2.34} & \textbf{-75.59} \\
         LED(2) &  \textbf{1.70}&           2.75  &  \textbf{2.34} & \textbf{-76.80} \\
      RBF(10,0) &         -3.28 & \textbf{-16.04} & \textbf{10.20} & \textbf{-89.14} \\
 RBF(10,0.0001) &         -2.60 & \textbf{-10.84} &  \textbf{7.30} & \textbf{-86.23} \\
  RBF(10,0.001) &         -0.72 & \textbf{-11.69} &  \textbf{7.59} & \textbf{-87.78} \\
      RBF(50,0) &         -3.28 & \textbf{-12.66} & \textbf{10.20} & \textbf{-88.31} \\
 RBF(50,0.0001) &         -0.27 & \textbf{-17.57} &  \textbf{7.63} & \textbf{-89.28} \\
  RBF(50,0.001) &         -0.15 & \textbf{-27.03} &  \textbf{2.36} & \textbf{-92.54} \\
        SEA(10) &         -0.09 &  \textbf{-7.61} &  \textbf{0.64} & \textbf{-90.07} \\
        SEA(20) &         -0.10 & \textbf{-14.12} &  \textbf{0.37} & \textbf{-90.46} \\
        airline &  \textbf{0.07}&  \textbf{-3.00} & \textbf{11.97} & \textbf{-83.08} \\
    electricity &  \textbf{3.32}&           2.78  &  \textbf{5.10} & \textbf{-67.75} \\
          poker &  \textbf{3.17}&           0.87  & \textbf{16.98} & \textbf{-76.80} \\
\midrule 
        Average & -0.22 &  -9.50 &  5.99 & -85.10 \\
\bottomrule
\end{tabular}